\documentclass[letterpaper, 10 pt, conference]{ieeeconf}  

\IEEEoverridecommandlockouts                              

\usepackage{amsmath,amsfonts}
\usepackage[ruled,vlined]{algorithm2e}
\usepackage{bbm} 
\usepackage{array}
\usepackage[caption=false,font=normalsize,labelfont=sf,textfont=sf]{subfig}
\usepackage{textcomp}
\usepackage{stfloats}
\usepackage{float}
\usepackage{url}
\usepackage{verbatim}
\usepackage{graphicx}
\usepackage{cite}
\usepackage{mathtools} 
\usepackage{subcaption}
\usepackage{amssymb}  

\usepackage{amsthm}
\usepackage{algpseudocode}
\usepackage{xcolor}
\usepackage{accents}
\usepackage{mathrsfs}
\usepackage[normalem]{ulem}

\def\beq{\begin{equation*}}
\def\eeq{\end{equation*}}
\def\bql{\begin{equation}}
\def\eql{\end{equation}}
\def\bqn{\begin{eqnarray*}}
\def\eqn{\end{eqnarray*}}
\def\bnl{\begin{eqnarray}}
\def\enl{\end{eqnarray}}

\usepackage{soul}
\usepackage{booktabs}
\usepackage{amsmath} 

\usepackage{multirow}

\usepackage[T1]{fontenc}
\usepackage[utf8]{inputenc}
\usepackage[skip=0.333\baselineskip]{caption}
\usepackage{siunitx} 
\newcolumntype{T}{S[table-format=3.3, input-symbols={()},
                    table-space-text-post={$^{***}$},
                    table-align-text-post=false]}
\usepackage[colorlinks=true, allcolors=black]{hyperref}
\usepackage{tabularx,booktabs}
\newcolumntype{C}{>{\centering\arraybackslash}X} 
\definecolor{green}{RGB}{11,155,13}

\SetCommentSty{mycommfont}

\SetKwRepeat{Do}{do}{while}

\DeclareMathOperator*{\argmin}{argmin}

\usepackage{yaacro}

\begin{acgroupdef}[list=acronimos]
    \acdef{JSG}{Joint State Graph}
\end{acgroupdef}

\newcommand{\nodeset}{\mathbb{V}}
\newcommand{\edgeset}{\mathbb{E}}

\title{\LARGE \bf
Heterogeneous Team Coordination on Partially Observable Graphs\\ with Realistic Communication
}

\author{Yanlin Zhou, Manshi Limbu, Xuan Wang, Daigo Shishika, and Xuesu Xiao
\thanks{All authors are with George Mason University {\tt\scriptsize \{yzhou30, klimbu2, xwang64, dshishik, xiao\}@gmu.edu}. 
}
}

\begin{document}
\maketitle
\thispagestyle{empty}
\pagestyle{empty}


\begin{abstract}
    Team Coordination on Graphs with Risky Edges (\textsc{tcgre}) is a recently proposed problem, in which robots find paths to their goals while considering possible coordination to reduce overall team cost. However, \textsc{tcgre} assumes that the \emph{entire} environment is available to a \emph{homogeneous} robot team with \emph{ubiquitous} communication. In this paper, we study an extended version of \textsc{tcgre}, called \textsc{hpr-tcgre}, with three relaxations: Heterogeneous robots, Partial observability, and Realistic communication. To this end, we form a new combinatorial optimization problem on top of \textsc{tcgre}. After analysis, we divide it into two sub-problems, one for robots moving individually, another for robots in groups, depending on their communication availability. Then, we develop an algorithm that exploits real-time partial maps to solve local shortest path(s) problems, with a A*-like sub-goal(s) assignment mechanism that explores potential coordination opportunities for global interests. Extensive experiments indicate that our algorithm is able to produce team coordination behaviors in order to reduce overall cost even with our three relaxations.\footnote{Full Version: \url{https://cs.gmu.edu/~xiao/papers/hpr-tcgre.pdf}} 

\end{abstract}

\section{Introduction}
\label{intro}
Team Coordination on Graphs with Risky Edges (\textsc{tcgre}) is a recently proposed problem~\cite{limbu2023team}, in which robots need to schedule  trajectories while considering possible coordination on the move to reduce overall team cost. 
Compared with the classical Multi-Agent Path Finding (\textsc{mapf}) problem~\cite{adler2002cooperative,hu2021distributed,yu2022surprising,surynek2010optimization,foerster2017stabilising,gupta2017cooperative, liu2021team}, where robots attempt to avoid collisions (and interactions) on their journeys, \textsc{tcgre} strives to exploit such opportunities, in terms of support between agents. This new problem possesses great 
practical value because it empowers robots with teamwork potential, which is especially useful in environments too complicated for a single robot to travel through, or in tasks so demanding that one robot may need assistance from others.

Most solutions to \textsc{tcgre} assume that the \emph{entire} environment is available to a \emph{homogeneous} robot team, with \emph{ubiquitous} communication. However, these assumptions may not hold in real-world scenarios. 
Robots may be deployed in unfamiliar environments and possess different capabilities to support or to be supported, while communication cannot happen anytime anywhere at no cost. 
Therefore, in this paper, we study an extended version of \textsc{tcgre}, called \textsc{hpr-tcgre}, which relaxes all three assumptions: 1) \textbf{H}eterogeneous team: different types of robots are involved, inducing a different edge cost, support cost, and risk reduction for each (pair of) robot type; 2) \textbf{P}artial observability: each robot can only observe the graph around it by sensors, while building up the entire graph in their memory as they move in the environment;  and 3) \textbf{R}ealistic communication: robots coordinate with one another via local communication channels to make coordination decisions and exchange environment knowledge only with their neighbors. 

We first complete a mathematical formulation of this problem, as a combinatorial optimization problem similar to \textsc{tcgre}~\cite{zhouIROS2024}. After analysis, we base our solution on dividing the problem into two sub-problems: 1) Individual movement: robots operate on their own when there are no nearby teammates; 2) Coordination: robots coordinate on the run when teammates are available. For the individual movement sub-problem, we then break it into two parts: i) each individual robot schedules a path to its sub-goal, solvable by any shortest path algorithm. ii) the robot decides a new sub-goal every time its partial map is updated, where A* can provide a good solution. Similarly, the coordination sub-problem is also separated into two parts: i) robots in the same partial map schedule their paths to their sub-goals, which is almost \textsc{tcgre}, solvable by algorithms for \textsc{tcgre}. ii) every robot in the same partial map decides a new sub-goal every time its partial map is updated, for which we develop a new algorithm that solves it as a multi-choice knapsack problem.
Extensive experimental results are presented and discussed to evaluate the performance of our proposed algorithm.

\begin{figure}
    \centering
    \includegraphics[width=0.99\linewidth]{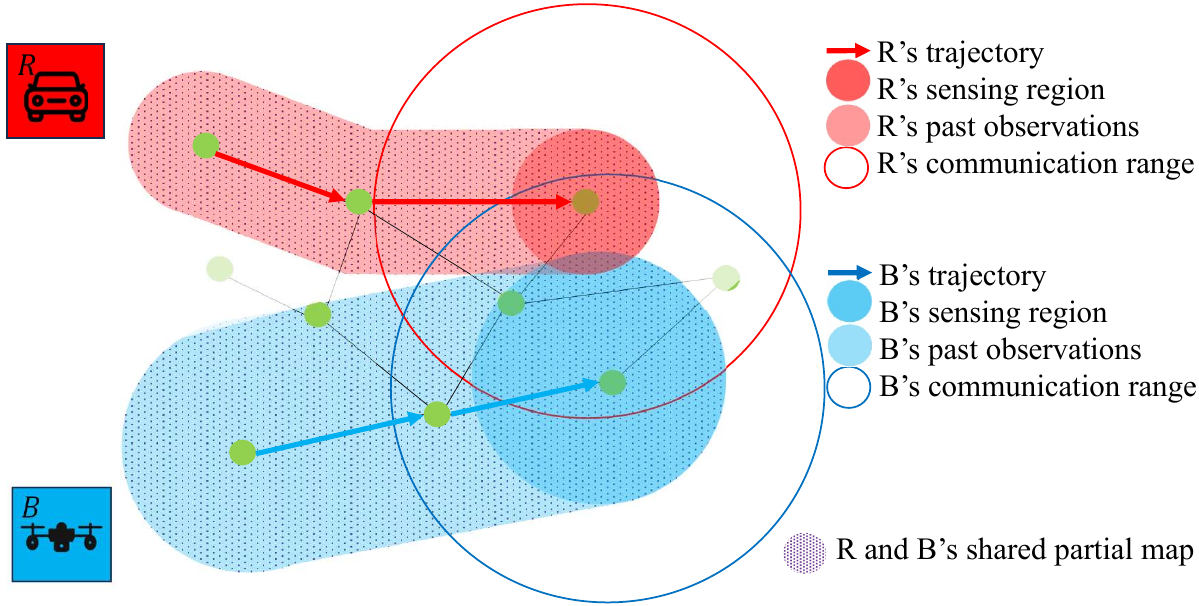}
     \vspace{1pt}
    \caption{An \textsc{hpr-tcgre} Example: A ground and an aerial robot gradually sense the environment as they traverse the graph, while exchanging knowledge and coordinating with each other when they are in communication range. }
    \label{fig1_HPRTCGRE}
    \vspace{10pt}
\end{figure}

\section{Related Work}
\label{sec::related}

The original \textsc{tcgre} problem has attracted some attention from researchers. Limbu et al.~\cite{limbu2023team} proposed a method that constructs a Joint State Graph (\textsc{jsg}) and converts the problem into a single-agent shortest path problem, solvable by Dijkstra's algorithm. They then developed the Critical Joint State Graph (\textsc{cjsg}) approach that removes many unnecessary edges in the \textsc{jsg} and mitigates the curse of dimensionality while maintaining optimality. To improve practicality of the solution so it works with more robots and larger graphs, Reinforcement Learning (RL)~\cite{limbuteam} is one way to scale up, but optimality is no longer a guarantee. Zhou et al.~\cite{zhouIROS2024} then reformulated the problem as a combinatorial optimization problem, elaborated on the importance of problem decomposition via mathematical analysis, and provided three classes of solutions: \textsc{jsg}-based approaches
are optimal but computationally expensive when the number of agents increases; coordination-based methods are optimal under an assumption that one coordination behavior only occurs for a limited number of times, therefore efficient in terms of the number of agents, but not the number of coordination behaviors; Receding-Horizon Optimistic Cooperative A* (\textsc{rhoc-a*}) search algorithms are efficient but sacrifice performance due to a limited horizon and limited robot pair choices.

However, the original \textsc{tcgre} problem is based on three strong assumptions: 1) All robots are the same; 2) The entire graph is known to the team a priori; and 
3) any team member can communicate with the central controller anytime anywhere at no cost.  
All these assumptions are not always realistic in real-world robot applications with limited perception range, different robot types, and constrained communication channels. Therefore,   the \textsc{hpr-tcgre} problem and its solution proposed in this work aim to relax all these three oversimplified assumptions.

\section{Problem Formulation}
\label{sec::method}
In the original \textsc{tcgre} problem, a team of $N$ homogeneous robots traverse an undirected graph \(\mathbb{G}=(\nodeset,\edgeset)\), where \(\nodeset\) is the set of nodes the robots can traverse to and \(\edgeset\) is the set of edges connecting the nodes, i.e., \(\edgeset\subset \nodeset\times \nodeset\). 
The team of robots traverse in the graph from their start nodes $\nodeset_0 \subset \nodeset$ to goal nodes $\nodeset_g \subset \nodeset$ via edges in \(\edgeset\). Each edge $e_{ij}=(V_i,V_j)\in \edgeset$ is associated with a cost. Specially, some edges with high costs are difficult to traverse through, denoted as risky edges $\edgeset' \subset \edgeset$, but with the support from a teammate from a supporting node, their costs can be significantly reduced. 
In this problem, such coordination behaviors only occur between two robots: one receiving robot receives support while traversing a risky edge, and another supporting robot offers support from some (nearby) location, called support node. Each risky edge $e_{ij} \in \edgeset'$ corresponds to certain support node(s) $\mathbb{S}_{e_{ij}} \subset \nodeset$ ($\mathbb{S}_{e_{ij}}=\emptyset$ if $e_{ij}\notin \edgeset'$). Additionally, the coordination also induces some cost for the supporter. The scheduling of all agents' movement and coordination relies on a central controller.

However, \textsc{tcgre} may not be applicable in many real-world cases as stated in Sec.~\ref{intro}, so in this paper we generalize it to \textsc{hpr-tcgre}, by relaxing three assumptions as follow:

1) Heterogeneous team: the team of robots involve $H$  types, who have different specialties possibly complementary to each other and thus contribute to better teamwork. As a result, an edge cost is not only dependent on the specific edge $e_{ij}$, but also the type of the robot traversing it $h_n$, 
represented as $c_{ij}^{h_n}$. For coordination, the reduced cost is related to the types of both robots conducting a coordination behavior $h_n$ and $h_m$, represented as $\tilde{c}_{ij}^{h_{nm}}$; 
the coordination cost is related to the type of the supporter $h_m$, denoted by $c'^{h_m}_{ij}$.

2) Partial observability: the environment is unknown in the beginning, and each robot can only use its sensors to observe the graph around it.  
That is to say, at time $t=0$, the graph is unknown to each robot $n$ in the team, i.e., $\mathbb{G}_n^0=(\mathbb{V}_n^{0}, \mathbb{E}_n^{0})$ where $\mathbb{V}_n^{0},\mathbb{E}_n^{0}=\emptyset$. Each robot $n$ can only sense a  region around its current node $l_n^t$; the range is dependent on the sensors of different robot types, i.e.,  $\mathbb{G}^h(l_n^t)$, with a memory of previously collected data. Therefore, each robot's knowledge of the graph, called partial map, is based on past and current observations $\mathbb{G}_n^t=\mathbb{G}_n^{t-1}\cup\mathbb{G}^h(l_n^t)$, as in Fig.~\ref{fig1_HPRTCGRE}. Moreover, each node corresponds to a set of coordinates, and each robot knows the coordinates of its goal, but not the path(s) to it, due to partial observability.

3) Realistic communication: without central control, robots have to rely on communication for coordination, with which they can also exchange their partial maps to facilitate navigation. Inspired by the 3-way handshake~\cite{hunt2002tcp}, we propose a similar process, where each robot $n$ broadcasts a hello message once at each time step; when another robot $m$ detects the message, it replies with another broadcast. If robot $n$ receives the response, it sends out a confirmation message. 
Then, the actual communication is established, and they start to coordinate after exchanging and combining their own partial maps and their coordination messages, also once per time step, i.e., $\mathbb{G}_n^{t},\mathbb{G}_m^{t}=\mathbb{G}_n^{t}\cup\mathbb{G}_m^{t}$
if $\mathscr{C}_{nm}^t=1$, as shown in Fig.~\ref{fig1_HPRTCGRE}. 
In this mechanism, the actual communication only starts when they get close, within the communication range---the minimum of the two transmission distances, i.e., $r_{nm}=\min(D_{nm},D_{mn})$. (The communication range in Fig.~\ref{fig1_HPRTCGRE} and Fig.~\ref{fig2_HPRTCGRE} are not following the strict definition, only for a better demonstration of the key ideas.)
Additionally, robots can communicate with each other via other intermediate robots and form ad-hoc networks~\cite{wu2004ad}, called robot groups. Denote any robot group that includes $n$ as ${\bf{RG}}_n^t$, where ${\bf{RG}}_n^t=\{n|\forall n \in {\bf{RG}}_n^t, \exists m \in {\bf{RG}}_n^t,~s.t.~d_{nm}^t\leq r_{nm}\}$ and ${\bf{RG}}_n^t={\bf{RG}}_n^t\cup{\bf{RG}}_m^t$ if $m\in{\bf{RG}}_n^t$. So for any robot pair in the same group, their communication is available, i.e., $\mathscr{C}_{nm}^t=1$ if $m\in{\bf{RG}}_n^t$, and $0$ otherwise.   
In this problem, we assume 
a constant cost for each communication,
which is trivial in the objective function. A more realistic communication cost model will be discussed in future work. 

Ultimately, each robot needs to schedule a path to reach their individual goal and coordinate with one another along the way, to minimize
the overall cost.

\subsection{Cost Model}
Similar to previous work~\cite{zhouIROS2024}, we can represent robots' decision making with two sets of 0/1 variables, a movement decision set $\mathcal{M}$ and coordination decision set $\mathcal{S}$. Specifically, without coordination, at each time step $t$,  each robot $n$ needs to decide where to go,  denoted as $\mathcal{M} = \{M_{ij}^{nt}| \forall i,j,\forall n, \forall t\}$, where $M_{ij}^{nt}=1$ represents robot $n$ traverses edge $e_{ij}$ at time $t$ and 0 otherwise. The general cost in this case is simply the cost of the chosen edge, i.e., $C_n^t = c_{ij}^{h(n)}$, where $M_{ij}^{nt} = 1$.
Staying is allowed for future coordination, i.e., $c_{ii}=0$. 

When coordination is possible --- after exchanging partial maps with other robots via communication, one robot can move to a support node to help another pass through an edge with reduced cost, coordinated via live communication --- each robot pair needs to decide whether they are going to coordinate, denoted as $\mathcal{S}=\{s_{nm}^t|\forall n,m, \forall t\}$. The cost is decided by both variables:
\begin{align}
\label{IndividualCost}
    C_n^t=\begin{cases}
        c_{ij}^{h_n}, ~~~~~~\textit{if} ~~ s_{nm}^t=~~0;\\
        \tilde{c}_{ij}^{h_{nm}},~~~~\textit{if} ~~ s_{nm}^t=~~1;\\
        c'^{h_m}_{ij},~~~~~\textit{if} ~~ s_{nm}^t=-1.
    \end{cases}
\end{align}
where $s_{nm}^t=1$ means robot $n$ receives support from $m$,  $s
_{nm}^t=-1$ indicates robot $n$ offers support to $m$, and $s_{nm}^t=0$ suggests no coordination between $n$ and $m$ at $t$. Such coordination only happens to two different robots, so $s_{nn}^t=0$. Additionally, coordination is only possible when robots know the existence of other robots by communication, i.e., $s_{nm}^t=0$ if $\mathscr{C}_{nm}^t=0$.

\vspace{-5pt}
\subsection{Problem Definition}
\label{definition}
\vspace{-5pt}
Given the node set $\nodeset$, the edge set $\edgeset$, support nodes for each edge $\mathbb{S}_{e_{ij}}$, $N$ robots with their starts $\nodeset_0$ and goals $ \nodeset_g$, their sensing regions $\mathbb{G}^h(\cdot)$ and communication range $\{r_{nm}\}$ (to determine $\mathscr{C}_{nm}^t$),
cost of each edge without and with coordination $c_{ij}^{h_n},\tilde{c}_{ij}^{h_{nm}}$, coordination cost ${c}'^{h_{nm}}_{ij}$, optimize the movement and coordination decisions $\mathcal{M}$ and $\mathcal{S}$, in order to minimize the total cost for each agent to traverse from its start to its goal within a time limit $T$.
Formally, the problem can be represented as
\begin{gather}
    \textcolor{black}{\min_{{\mathcal{M}},{\mathcal{S}}} \sum_{t=0}^{T-1} \sum_{n=1}^N C_{n}^t;} \label{objective}\\ \allowdisplaybreaks[4]
    s.t.~~
     \sum_{\forall m \in \{1,2,...,N\}}|s_{nm}^t|\leq 1,\nonumber\\ \allowdisplaybreaks[4]
    \forall n\in \{1,2,...,N\}, \forall t\in \{0,1,...,T-1\}. \label{constraint1:supportnum}\\ \allowdisplaybreaks[4]
    s_{nm}^t, s_{mn}^t\in\{-1, 0,1\}, s_{nm}^t+ s_{mn}^t=0, \nonumber\\ \allowdisplaybreaks[4]
    \forall n,m\in \{1,2,...,N\}, \forall t\in \{0,1,...,T-1\}. \label{constraint2:supportpair}\\ \allowdisplaybreaks[4]
    \textcolor{black}{s_{nm}^t=0,~\textit{if}~\mathscr{C}_{nm}^t=0.} \label{constraint3:communicationrange}\\ \allowdisplaybreaks[4]
    \textcolor{black}{\mathbb{G}_n^{t}=\mathbb{G}_n^{t-1}\cup\mathbb{G}^h(l_n^t)
    .}\label{constraint4:sensingregion}\\ 
    \allowdisplaybreaks[4]
    \textcolor{black}{\mathbb{G}_n^{t}, \mathbb{G}_m^{t}=\mathbb{G}_n^{t}\cup\mathbb{G}_m^{t},~\textit{if}~\mathscr{C}_{nm}^t=1
    .} \label{constraint4:sharedsensingregion}\\\allowdisplaybreaks[4]
    l_n^0=\nodeset_0(n), l_n^T=\nodeset_g(n), \forall n\in \{1,2,...,N\}.\label{constraint5:startsandgoals}\\ \allowdisplaybreaks[4]
    \sum_{\forall e_{ij}\in{\mathcal{N}_{l_n^t}}}M_{ij}^{nt}=1,  \sum_{\forall e_{ij}\notin{\mathcal{N}_{l_n^t}}}M_{ij}^{nt}=0,                   \nonumber\\ \allowdisplaybreaks[4]
    \forall n\in \{1,2,...,N\}, \forall t\in \{0,1,...,T-1\}.\label{constraint7:movenum}\\ \allowdisplaybreaks[4]
    \sum_{\forall n}\sum_{\forall i\neq j} M_{ij}^{nt}\neq0,  
    \forall t\in\{0,1,...,T-1\}.\label{constraint8:correlation}
\end{gather}
Eqn.~\eqref{objective} suggests the goal of the problem is to minimize the total cost of all agents across all time steps with two decision variables, movement set $\mathcal{M}$ and coordination set $\mathcal{S}$. Eqn.~\eqref{constraint1:supportnum} enforces, at each time step, each robot can participate in at most one coordination behavior. Eqn.~\eqref{constraint2:supportpair} regulates that, at each time step, one coordination behavior only occurs between one robot pair. 
Eqn.~\eqref{constraint3:communicationrange} shows that coordination is only possible when communication is available.
Eqn.~\eqref{constraint4:sensingregion} defines the update mechanism of a partial map.
Eqn.~\eqref{constraint4:sharedsensingregion} is for graph knowledge exchange when communication is available. 
Eqn.~\eqref{constraint5:startsandgoals} set the start and the goal for each robot. Eqn.~\eqref{constraint7:movenum} guarantees, at each time step, a robot can only move to a neighbor node or stay still in the neighbor set $\mathcal{N}_{l_n^t}$ of node $l_n^t$. Eqn.~\eqref{constraint8:correlation} assures no unnecessary stagnation. 
Notice that the sensing region $\mathbb{G}^h(\cdot)$ and communication range $\{r_{nm}\}$ are 
 decided by hardware, so we treat them as given inputs in the problem.

\section{Mathematical Analysis}
\label{sec::sol}
In this section, we show the NP-hardness and present problem analysis of \textsc{hpr-tcgre}. 

\vspace{-5pt}
\subsection{NP-Hardness}
\vspace{-5pt}
By constraining the number of robot types to one, and the sensing region of any robot and communication range between any robot pairs to be large enough to cover the entire graph, the \textsc{hpr-tcgre} problem becomes exactly \textsc{tcgre}. That is to say, \textsc{tcgre} is a specific version of \textsc{hpr-tcgre}; since TCGRE is already
proven NP-hard~\cite{zhouIROS2024}, \textsc{hpr-tcgre} is also NP-hard.

\vspace{-5pt}
\subsection{Problem Analysis}
\vspace{-5pt}
\subsubsection{Cost Model}
Similar to previous work~\cite{zhouIROS2024}, since the coordination only occurs in a pair, then the supporting cost can be moved to the receiving robot, as we only care about the overall cost, so Eqn.~(\ref{IndividualCost}) becomes
\begin{align}
\label{IndividualCost2}
    C_n^t=\begin{cases}
        c_{ij}^{h_{n}}, ~~~~~~\textit{if} ~~ s_{nm}^t=~~0;\\
        \hat{c}_{ij}^{h_{nm}},~~~~\textit{if} ~~ s_{nm}^t=~~1;\\
        0,~~~~~~~~\textit{if} ~~ s_{nm}^t=-1,
    \end{cases}
\end{align}
where $\hat{c}_{ij}^{h_{nm}}=\tilde{c}_{ij}^{h_{nm}}+c'^{h_m}_{ij}$.

\subsubsection{Distributed Goals}
Our fundamental goal is to utilize communication and coordination among robots to reduce the overall cost of all robots for reaching their goals.
However, only when the communication channel is established, will there be any opportunity to  coordinate, so the cost is only possible to be reduced with an active channel (Eqn.~\eqref{constraint3:communicationrange}). Thus, the objective function can be rewritten as

\begin{gather}
    \min_{{\mathcal{M}},{\mathcal{S}}} \sum_{t=0}^{T-1} \sum_{n=1}^N (1-\mathscr{C}_{nm}^t) c_{ij}^{h_{n}} \label{objective1:nocommunication}\\
    +\sum_{t=0}^{T-1} \sum_{n=1}^{N}\mathscr{C}_{nm}^t[ (1-s_{nm}^t)c_{ij}^{h_{n}}+s_{nm}^t \hat{c}_{ij}^{h_{nm}}],  \label{objective2:communication}
\end{gather}
where Eqn.~(\ref{objective1:nocommunication}) refers to independent operation of each individual robot when no teammates are detected, while Eqn.~(\ref{objective2:communication}) involves communication and coordination.
\section{Solution}
\label{sec::solutions}
Based on the mathematical analysis, the decision variable $s_{nm}^t$ is only in Eqn.~(\ref{objective2:communication}) but not Eqn.~(\ref{objective1:nocommunication}), one simple solution is to divide the problem into two sub-problems: independent movement   (Eqn.~(\ref{objective1:nocommunication})) and coordination (Eqn.~(\ref{objective2:communication})), depending on whether other robots are detected. Regarding detection, since the communication range is usually much larger than the sensing range, we use communication as the indicator, i.e., $\mathscr{C}_{nm}^t=0/1$.

\begin{figure*}
    \centering
    \includegraphics[width=0.9\linewidth]{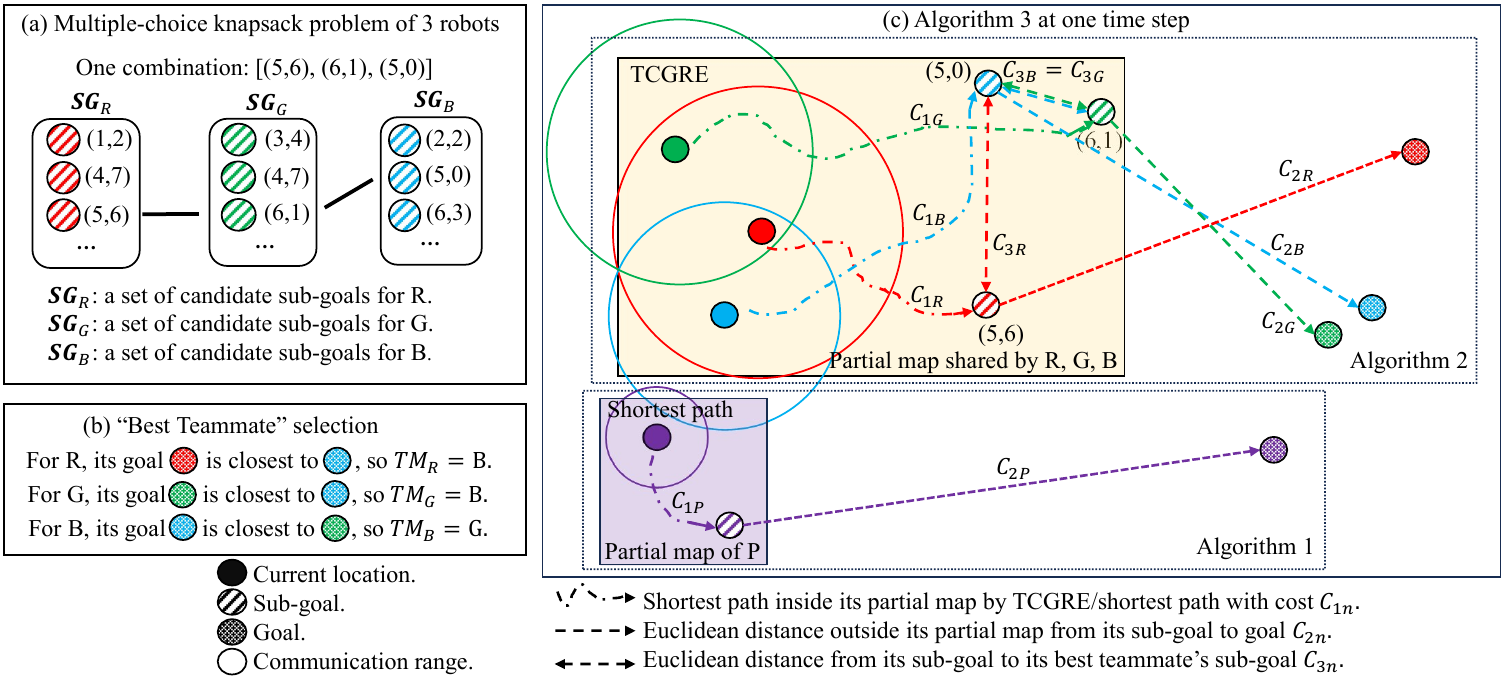}
    \vspace{3pt}
    \caption{Solution Overview.}
    \label{fig2_HPRTCGRE}
\end{figure*}

\subsection{Individual Movement}
\label{individualmovement}
When a robot detects no teammates, i.e., $\mathscr{C}_{nm}^t=0, \forall m$, no coordination is available, so it approaches its goal individually.
The objective of all robots in Eqn.~(\ref{objective1:nocommunication}) can be distributed to each individual robot $n$ as
\begin{gather}
    \min_{{\mathcal{M}}} \sum_{t=0}^{T-1}  c_{ij}^{h_n}. \label{objective3:individualmovement}
\end{gather}
If the entire graph is provided, it is a shortest path problem.
However, partial observability (Eqn.~\eqref{constraint4:sensingregion}) makes the nodes and edges outside its partial map $\mathbb{G}_n^t$ invisible. Hence, the shortest path algorithm is only applicable locally inside the partial map, outside which we use Euclidean distance to estimate the real cost. Specifically, to run the shortest path algorithm, we need to assign a goal for each robot in its partial map, called sub-goal. Moreover, the partial map grows as it moves, so the sub-goal also changes constantly along its way. 
The idea is \textit{if all the sub-goals selected for an individual robot are on its optimal path to its final goal, the solution is optimal}. That is to say, this sub-problem can be equivalently transformed into deciding a sub-goal for each partial map.

In this paper we propose an A*-like algorithm, as in Alg.~\ref{alg1}, that decides a sub-goal, using the sum of 1) $C_1$: its minimal cost from its current node to a potential sub-goal (line 5), and 2) $C_2$: the Euclidean distance between this sub-goal and the final goal (line 6). If the Euclidean length of an edge is no greater than its cost, this heuristic is admissible, as the first minimal cost is optimal and the second Euclidean distance is an underestimate of the minimum cost between the sub-goal and final goal. 

\begin{algorithm}    \caption{$Individual$ ($\mathbb{G}_n^t, l_n^t, \nodeset_g[n]$)}
    \label{alg1}
    \nl Generate a candidate sub-goal list, ${\bf{SG}}_n^t$, in $\mathbb{G}_n^t$;\\
    \nl \bf{if} $\nodeset_g[n]\in\mathbb{G}_n^t$, \bf{then} ${\mathbf{SG}}_n^t=[\nodeset_g[n]]$;\\ 
    \nl $min\_cost\leftarrow\infty$; $min\_{\mathbf{Solution}}\leftarrow\emptyset$; \\
    \nl \For{$subgoal\in{\bf{SG}}_n^t$}{
        \nl ${\mathbf{P}}, C_1=ShortestPath(\mathbb{G}_n^t,l_n^t,subgoal,\mathbf{c})$;\\
        \nl $C_2=d(subgoal,\nodeset_g[n])$;\\
        \nl \If{$C_1+C_2<min\_cost$}{
          \nl  $min\_cost=C_1+C_2$;\\
          \nl  $min\_{\mathbf{Solution}}=[{\mathbf{P}}, C_1]$;
        } 
    }
\nl\Return{$min\_{\mathbf{Solution}}$}
\end{algorithm}


\subsection{Coordination}
While robot $n$ detects some teammates, i.e., $\mathscr{C}_{nm}^t=1, \exists m$, directly or indirectly via intermediate robots, they need to consider coordination to minimize their total cost,
almost equivalent to \textsc{tcgre}, except with heterogeneous robots. 
In a robot group $\bf{RG}$, all robots communicate and share partial maps. 
Similar to Sec.~\ref{individualmovement}, the problem becomes deciding sub-goals for each $\bf{RG}$. In addition, they also need to consider future coordination opportunities; the main idea is that,
\textit{for any robots in a robot group whose goals are close, they are potentially good teammates, and thus their sub-goals should also be close.}
In other words, in Alg.~\ref{alg2}, (in line 12) we approximate Eqn.~(\ref{objective2:communication}) with
\begin{gather}
    \sum_{t=0}^{T-1} \sum_{n=1}^{N} C_{1n} + C_{2n} + \alpha_n C_{3n}, \label{objective4:communication}
\end{gather}
where $\sum_{t=0}^{T-1} \sum_{n=1}^{N}C_{1n}$ is Eqn.~(\ref{objective2:communication}) inside a partial map, solvable by any algorithms for \textsc{tcgre}~\cite{zhouIROS2024}, with one more dimension (robot type) in edge costs (lines 6 and 8); $C_{2n}$ is the Euclidean distance between each robot $n$'s sub-goal and final goal (line 9); $C_{3n}$ is the sub-goal distance between each robot $n$ and its best teammate $TM_n$ (line 11), whose final goal is closest to robot $n$'s (line 10),  
to encourage robots whose final goals are close to each other to stay close along their journey so they have high chances of supporting each other to reduce cost. $C_{3n}$ is weighed by a coefficient $\alpha_n$ with respect to $C_{1n}$ and $C_{2n}$. Note that ${\bf{L}}_{\mathbf{RG}}^t=\{l_n^t|\forall n \in \bf{RG}\}$; ${\bf{c}}=\{c_{ij}^{h_n}|\forall e_{ij}\in\mathbb{G}_{\bf{RG}}^t, \forall h_n\}$; $\hat{\bf{c}}=\{\hat{c}_{ij}^{h_{nm}}|\forall e_{ij}\in\mathbb{G}_{\bf{RG}}^t, \forall h_n, h_m\}$.

We treat Eqn.~\eqref{objective4:communication} for each $\bf{RG}$ as a multiple-choice knapsack problem~\cite{kellerer2004multiple} with special weights as in Fig.~\ref{fig2_HPRTCGRE}, where we need exactly one item (sub-goal) from each class (candidate sub-goals of each robot) of a k-class knapsack, with a cost calculated by Eqn.~\eqref{objective4:communication}. Minimizing cost can be regarded as maximizing cost reduction, similar to previous work~\cite{zhouIROS2024}. For this problem, we use Cartesian product to produce all sub-goal combinations (line 3) and run an exhaustive search (line 4) to find the best combination (lines 12-14). 


\begin{algorithm}    \caption{$Coordination$ ($\mathbb{G}_{\mathbf{RG}}^t, {\bf{L}}_{\mathbf{RG}}^t, \mathbf{RG}, \nodeset_g$)}
    \label{alg2}
    \nl Generate a list of candidate sub-goal lists, ${\mathbb{SG}}^t=\{{\mathbf{SG}}_n^t,\forall n\in\mathbf{RG}\}$, in $\mathbb{G}_{\mathbf{RG}}^t$;\\
    \nl \bf{if} $\nodeset_g[n]\in\mathbb{G}_{\mathbf{RG}}^t$, \bf{then} ${\mathbf{SG}}_n^t=[\nodeset_g[n]]$;\\
    \nl $min\_cost\leftarrow\infty$;~$min\_{\bf{Solution}}\leftarrow\emptyset$;\\
    \nl ${\mathbf{SSG}}^t= Product(*{\mathbb{SG}}^t)$\\
    \nl \For{${\bf{SG}}^t\in{\mathbf{SSG}}^t$}{
    \nl ${\mathbb{P}},\mathbf{C}_{1}=\textsc{tcgre}(\mathbb{G}_{\mathbf{RG}}^t,{\bf{L}}_{\mathbf{RG}}^t,{\bf{SG}}^t,\mathbf{c},\hat{\mathbf{c}})$;\\
    \nl \For{$n\in\mathbf{RG}$}{
        \nl $C_{1n}=\mathbf{C}_{1}[n]$;\\
        \nl  $C_{2n}=d({\bf{SG}}^t[n],\nodeset_g[n])$;\\
        \nl $TM_n=\argmin_{m\in\mathbf{RG}} d(\nodeset_g[n],\nodeset_g[m])$;\\
        \nl $C_{3n}=d({\bf{SG}}^t[n], {\bf{SG}}^t[TM_n])$;
    }
    \nl $C=\sum_{\forall n\in \mathbf{RG}} C_{1n}+C_{2n}+\alpha_n C_{3n}$;\\
    \nl \If{$C<min\_cost$}{
        \nl  $min\_cost=C$;\\
        \nl  $min\_{\bf{Solution}}=[{\mathbb{P}},\mathbf{C}_{1}$];\\
    } 
    }
\nl\Return{$min\_{\bf{Solution}}$}
\end{algorithm}


\subsection{Overall Solution}
Since we already have Alg.~\ref{alg1} and Alg.~\ref{alg2} for the two sub-problems, now we only need to apply them in the corresponding scenarios.
The overall algorithm is shown in Alg.~\ref{alg3}. For each time step, we search each robot’s communication channel and build robot groups $\mathbb{RG}^t$: one robot group $\bf{RG}$ includes a group of robots who are communicating directly or indirectly (sharing the same partial map). For each $\bf{RG}$, if it only has one robot, it means the robot is moving individually and we run Alg.~\ref{alg1} (lines 8-15); otherwise, the robot group may need coordination and we run Alg.~\ref{alg2} (lines 16-25). 
Note that in Alg.~\ref{alg3}, we build $\bf{RG}$ for individuals too ($|\bf{RG}|=1$) for code efficiency; in concept individuals don't form $\bf{RG}$s.

\begin{algorithm}    \caption{\textsc{hpr-tcgre} ($\mathbb{G},  \nodeset_0, \nodeset_g$)}
    \label{alg3}
    \nl Initialize paths for all robots $\mathbb{P}_{min}=[\emptyset]*N$;\\
    \nl Initialize total cost $total\_cost = 0$; \\ 
    \nl Initialize every robot's partial map $\mathbb{G}_n^0=\emptyset, \forall n$;\\
    \nl Initialize the list of all robot groups $\mathbb{RG}^0=\emptyset$;\\
    \nl \For {$t=1$ to $T-1$}{
    \nl Update $\mathbb{RG}^t$ and partial maps $\mathbb{G}_n^t,\forall n$;\\
    \nl \For{$\mathbf{RG}\in\mathbb{RG}^t$}{
    \nl \uIf{$|\mathbf{RG}|=1$}{
       \nl ${\mathbf{P}}, {C} = Individual (\mathbb{G}_n^t, l_n^t, \nodeset_g[n])$;\\
       \nl $edge = {\mathbf{P}}[0]$; $cost=$cost of $edge$; \\
       \nl $\mathbb{P}_{min}[n].append(edge)$;\\
       \nl $total\_cost=total\_cost+cost$;\\       
       \nl Update its coordinate $l_n^t=edge[1]$;\\
         \nl \If{$l_n^t=\nodeset_g[n]$}{\nl Ignore $n$ while updating $\mathbb{RG}^{t+1}$;}
    }
    \nl \Else{
       \nl Share partial maps $\mathbb{G}_{\mathbf{RG}}^t=\bigcup_{\forall n \in\mathbf{RG}}\mathbb{G}_n^t$;\\
            \nl  ${\mathbb{P}},\mathbf{C}=Coordination$ ($\mathbb{G}_{\mathbf{RG}}^t, {\bf{L}}_{\mathbf{RG}}^t, \mathbf{RG}, \nodeset_g$);\\
       \nl \For {$n\in \mathbf{RG}$}{
         \nl $edge = {\mathbb{P}}[n][0]$; $cost=$cost of $edge$;\\
         \nl $\mathbb{P}_{min}[n].append(edge)$;\\
         \nl $total\_cost=total\_cost+cost$;\\ 
         \nl Update its coordinate $l_n^t=edge[1]$;\\
         \nl \If{$l_n^t=\nodeset_g[n]$}{
         \nl Ignore $n$ while updating $\mathbb{RG}^{t+1}$;
         }
       }        
    }
    }
    }
\nl\Return{$\mathbb{P}_{min},total\_cost$}
\end{algorithm}

Specifically, to better demonstrate the key ideas of the algorithm, some details are not shown in the algorithm definitions, which have great impact on the algorithm performance and efficiency:
1) In both Alg.~\ref{alg1} and Alg.~\ref{alg2}, the candidate sub-goals are only nodes on the frontier of each partial map that are not dead ends. To improve efficiency, candidates may be limited to nodes near its current node, at the expense of some performance loss.
2) In Alg.~\ref{alg2}, while calculating $\sum_n C_{3n}$ for all combinations of one $\bf{RG}$, we only need to sum the counterpart for each cluster, without going through every combination; a cluster is a minimal set of robots that includes all "best teammates" of the robots in the cluster.
3) In Alg.~\ref{alg3}, Alg.~\ref{alg1} or Alg.~\ref{alg2} only needs to be called when the partial map is updated or some members of the $\bf{RG}$ are changed. 4) In both Alg.~\ref{alg1} and Alg.~\ref{alg2}, when deciding sub-goal(s), an $\epsilon-$greedy model~\cite{sutton2018reinforcement} can be applied to escape local optimums: we select the "best" sub-goal(s) with a probability $\epsilon$, and other candidates with an even chance.

\section{Results}
\label{sec::results}
\vspace{-3pt}
We conduct experiments on a variety of large graphs of the same size but different structures to evaluate the effect of number of robot types, size of sensing region, and size of communication range on the algorithm performance. We also conduct ablation experiments to evaluate each part of the algorithm.

\subsection{Experiment Settings}
 We vary the number of robot types (labeled "$\#$" in the results), sensing factor (the factor multiplying the graph length is the sensing distance, within which the nodes and edges can be detected), and communication factor (similar to sensing factor), which represent the three relaxations we have in this problem. The two sensing/communication factors are common for every robot; the common factors need to multiply another unique coefficient for each robot type to simulate the definition of an edge cost in the problem.
 For each setting, we run the algorithms in 10 graphs. Every graph is randomly generated with 20 nodes, 50\% edges, among which 20\% are risky edges,  and random starts and goals for 7 robots. 
 Since \textsc{hpr-tcgre} is a new and complex problem, there is no state-of-the-art algorithm to compare against, and we conduct an ablation experiment with 
 the following baselines: 1) The naive approach where every robot is moving individually without coordination (labeled "Naive"). 2) Alg.~\ref{alg3} without consideration of future coordination (labeled "Alg.~\ref{alg3} w/o. $C_3$"). 3) Alg.~\ref{alg3} with an epsilon-greedy model while deciding sub-goal(s) (labeled "w/. epsilon").
 Furthermore, because we are experimenting on large graphs with a lot of risky edges, we are only using \textsc{rhoca}~\cite{zhouIROS2024} for \textsc{tcgre} in Alg.~\ref{alg2}.

\begin{figure*}
    \begin{minipage}[t]{0.49\linewidth}
    \centering
    \includegraphics[width=0.8\textwidth]{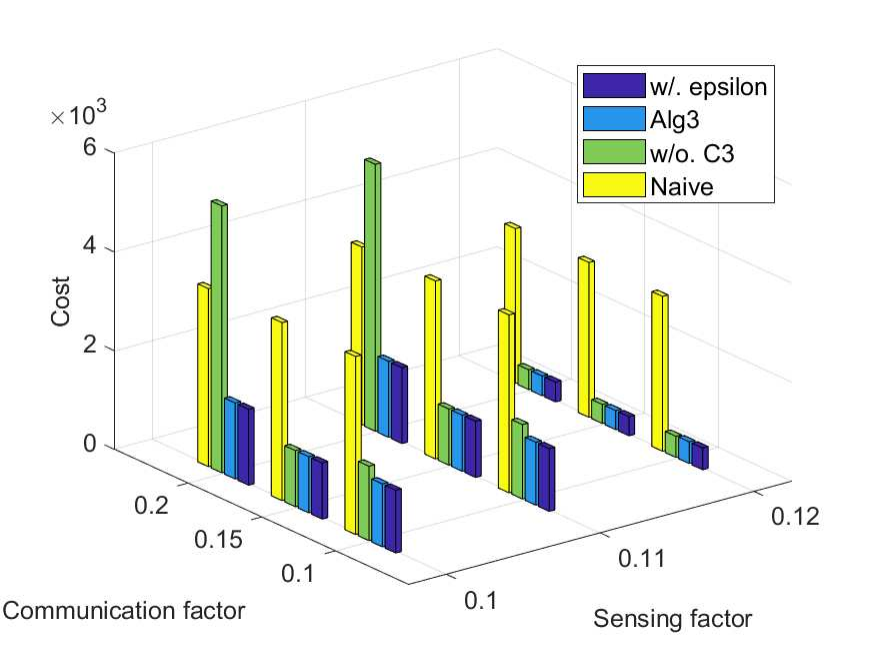}
    \caption{Cost.}
    \label{fig_cost}
    \end{minipage}
    \hfill
    \begin{minipage}[t]{0.5\linewidth}
    \centering
    \includegraphics[width=0.8\textwidth]{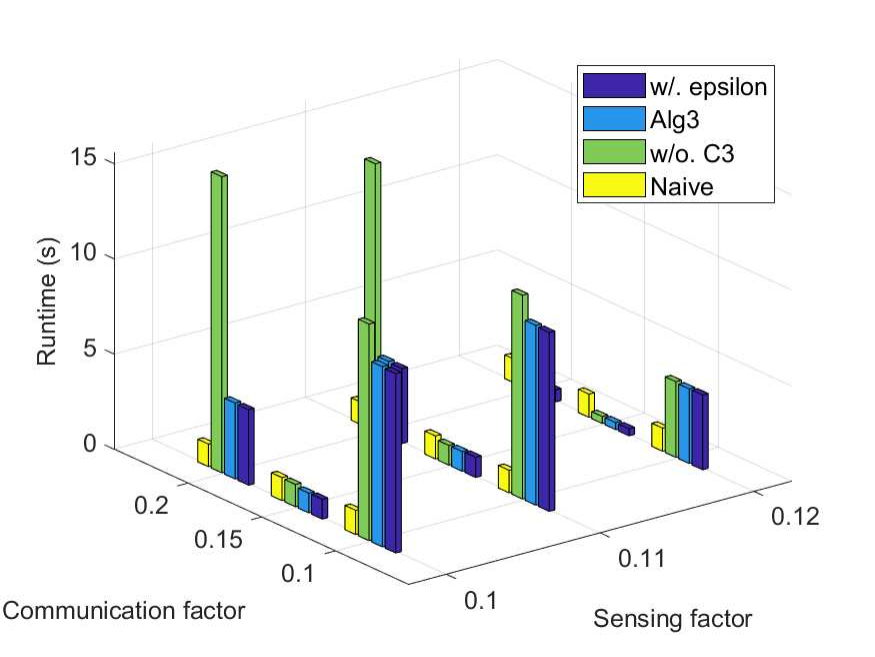}
    \caption{Runtime.}
    \label{fig_runtime}
    \end{minipage}
\end{figure*}

\subsection{Results}
Theoretically, the runtime does not  change with the number of robot types, because the computation time to construct and search on \textsc{JSG}\cite{limbu2023team} for \textsc{tcgre} remains the same. \textsc{JSG} is not implemented because the large size of our graphs. 
Therefore, we do not show results with different numbers of robot types in the figures.   

In Fig.~\ref{fig_cost}, the naive algorithm consistently produces high cost, due to dead ends, the node(s) that trap robots locally due to the over-optimistic heuristic of A* algorithms. This problem also significantly affects other algorithms, but their coordination can sometimes take them out of the local optimums of their individual paths. However, this is not always the case, e.g., the two high-cost bars for the baseline without C3, as a result of the randomness of the graph. Specifically, the epsilon-greedy model spends some efforts on exploring the graph, rather than exploiting the partial map information greedily. Though it takes some cost in exploration, it prevents the robots from the traps of the dead ends. 

As shown in Fig.~\ref{fig_runtime}, the two high bars of the algorithm without C3 is caused by the dead ends, the same reason mentioned above.  Counterintuitively, the runtime of the naive approach is always low, even though it encounters dead ends as in Fig.~\ref{fig_cost}. However, because the algorithm is simple without coordination and thus fast for each time step, so even if it terminates due to the time limit $T$, the total cost of all time steps is still low.

In summary, the algorithm performance is affected by the graph structure; so is the coordination. Therefore, in the future, we plan to explore how exactly will the structure affect the coordination.

\section{Conclusions and Discussions}
\label{sec::conclusions}
We study a more general version of \textsc{tcgre}~\cite{zhouIROS2024}, called \textsc{hpr-tcgre}, by relaxing three assumptions with Heterogeneous robots, Partial observability, and Realistic communication. We construct a new combinatorial optimization problem based on the previous one~\cite{zhouIROS2024}. 
After analysis, we break down the problem into two sub-problems: individual movement and coordination. The coordination sub-problem is then broken into \textsc{tcgre} problems inside their shared local partial maps of robot groups. Globally for the group we solve the multi-choice knapsack problem with an estimated cost model considering future coordination. 


Our \textsc{hpr-tcgre} problem is difficult as the robots are moving in an unknown environment without prior knowledge about the environment in the beginning. We apply a communication model to help them explore the graph and enable coordination. But the model is not completely realistic, because we are not considering the transmission cost function and the transmission time; both involve how much graph information they need to exchange at once and at what distances they should exchange the information, which may also introduce a data filtering mechanism to decide which nodes and edges are worth going to. Furthermore, we use an estimated cost model that considers future coordination, because robots fall into individual moving mode and coordination mode interchangeably, and they are not really two independent sub-problems. However, the estimation may not work very well in complex environments, as it is a simple linear model. In the future, it may be necessary to develop a teammate position prediction mechanism, with which the problem can be solved in a more comprehensive manner.

\newpage
\bibliographystyle{ieeetr}
\bibliography{bib}
\end{document}